\pdfoutput=1
\documentclass[11pt]{article}

\usepackage{emnlp2021}

\usepackage{times}
\usepackage{latexsym}

\usepackage{graphicx}
\usepackage{amsmath}
\usepackage{bm}
\usepackage{multirow}
\usepackage{amssymb}

\usepackage[T1]{fontenc}
\usepackage[utf8]{inputenc}

\usepackage{microtype}

\title{Transductive Learning for Unsupervised Text Style Transfer}

\author{
    Fei Xiao$^{1,4}$, 
    Liang Pang$^{1*}$, 
    Yanyan Lan$^{3*}$, 
    Yan Wang$^{5}$, 
    Huawei Shen$^{1,4}$, 
    Xueqi Cheng$^{2,4}$ \\
    $^{1}$Data Intelligence System Research Center\\
    and $^{2}$CAS Key Lab of Network Data Science and Technology, \\
    Institute of Computing Technology, Chinese Academy of Sciences \\
    $^3$Institute for AI Industry Research, Tsinghua University \\
    $^4$University of Chinese Academy of Sciences \, \, \, 
    $^5$Tencent AI Lab \\
    \texttt{\{xiaofei19s, pangliang, shenhuawei, cxq\}@ict.ac.cn} \\
    \texttt{lanyanyan@tsinghua.edu.cn},\, \texttt{brandenwang@tencent.com}
}

\begin{document}
\maketitle
\begin{abstract}
Unsupervised style transfer models are mainly based on an inductive learning approach, which represents the style as embeddings, decoder parameters, or discriminator parameters and directly applies these general rules to the test cases. However, the lacking of parallel corpus hinders the ability of these inductive learning methods on this task. As a result, it is likely to cause severe inconsistent style expressions, like {\em the salad is rude}.
To tackle this problem, we propose a novel transductive learning approach in this paper, based on a retrieval-based context-aware style representation. Specifically, an attentional encoder-decoder with a retriever framework is utilized. It involves top-$K$ relevant sentences in the target style in the transfer process. In this way, we can learn a context-aware style embedding to alleviate the above inconsistency problem. In this paper, both sparse (BM25) and dense retrieval functions (MIPS) are used, and two objective functions are designed to facilitate joint learning. Experimental results show that our method outperforms several strong baselines. The proposed transductive learning approach is general and effective to the task of unsupervised style transfer, and we will apply it to the other two typical methods in the future.
\let\thefootnote\relax\footnotetext{*Corresponding Author}
\end{abstract}

\section{Introduction}
Text style transfer is an essential topic of natural language generation, which is widely used in many tasks such as sentiment transfer~\citep{hu2017toward, shen2017style}, dialogue generation~\citep{zhou2018emotional, niu2018polite, su2021prototype}, and text formalization~\citep{jain2019unsupervised}. The target is to change the style of the text while retaining style-independent content. As it is usually hard to obtain large parallel corpora with the same content and different styles, unsupervised text style transfer becomes a hot yet challenging research topic in recent years.

\begin{figure}
\centering %使插入的图片居中显示
\includegraphics[scale=0.46]{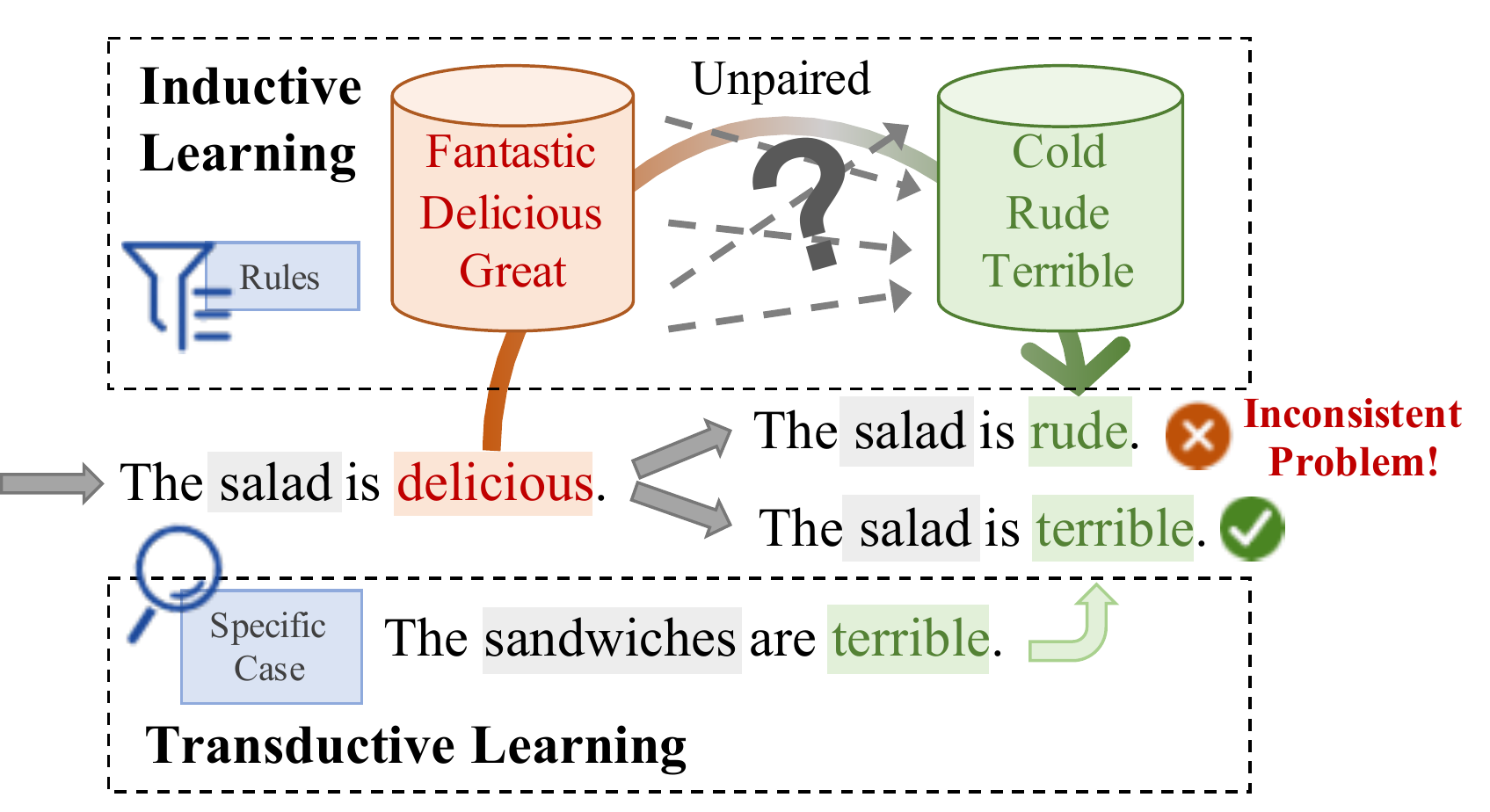}
\caption{\label{fig:introduction} Illustration of the inconsistency problem and the idea of our transductive learning approach.} %插入图片的标题，一般放在图片的下方，放在表格的上方
\vspace{-15pt}
\end{figure}

Most existing methods in this area try to find the general style transfer rules with an inductive learning paradigm, where style is represented as a specific form, e.g., embeddings, decoder parameters, or classifier parameters. 
For example, embedding based methods~\citep{shen2017style, lample2018multiple, john2019disentangled, dai2019style, yi2020text, he2020probabilistic} utilize a highly generalized style embedding to replace the original sentence style and direct the generation process. Decoder based methods~\citep{prabhumoye2018style, fu2018style, luo2019dual, gong2019reinforcement, krishna2020reformulating} use multiple decoders for generation, where each encoder corresponds to an independent style. Classifier based methods~\citep{wang2019controllable, liu2020revision, mai2020plug} employ the gradient of a pre-trained style classifier to edit the latent representation of the target text.

It has been well accepted that inductive learning methods have the ability to work well when there are numerous supervised labels. However, in the case of unsupervised style transfer, we are just given corpora with different styles without knowing the parallel relation, i.e.~supervision label for this task. As a result, the inductive learning methods fail to produce an accurate style transfer rule, leading to generating some severe inconsistent texts, such as `\emph{the salad is rude}', as shown in Figure~\ref{fig:introduction}. The underlying reason for this phenomenon is that a perfect style is usually highly dependent on the context, e.g.~`\emph{terrible}' for `\emph{the salad}' and '\emph{rude}' for '\emph{a person}'. Without a large scale of parallel data, it is difficult to learn a general style transfer rule working for various contexts.

Inspired by the idea of transductive learning~\citep{Vapnik1998} and some successful historical examples, such as Transductive SVM~\citep{joachims1999transductive}, we propose to introduce transductive learning to the area of unsupervised text style transfer. Specifically, transduction learning reasoning from specific cases to specific cases, which avoids learning a general rule to represent the style. For example, once we get a reference sentence `\emph{the sandwiches are terrible}' with negative emotion, a transductive learning method may connect the two sentences by the two kinds of food, e.g. `\emph{salad}' and `\emph{sandwiches}', then use `\emph{terrible}' to express the negative emotion for the food '\emph{salad}'.

From the above discussion, we can see that there are two challenges in applying transductive learning to unsupervised text style transfer: 1) how to find specific samples that are beneficial for the style transfer of the current text; 2) how to use the style expressions in these samples to complete the style transfer process. To tackle these two challenges, we propose a novel \textbf{T}ran\textbf{S}ductive \textbf{S}tyle \textbf{T}ransfer (TSST) model. In TSST, a retriever is employed to obtain the required similar samples, which tackles the first challenge. An attention-based encoder-decoder framework is then utilized to combine the specific samples to tackle the second challenge. Specifically, TSST first encodes the original text to a contextual representation and a style-independent embedding. Then either sparse (BM25) or dense retrieval functions (MIPS) are used to find the top-$K$ samples in the target style corpus, which are encoded by the same encoder. After that, a recurrent decoder is utilized to generate transfer text word by word based on the representation of those retrieved samples, contextual representation, and the representations in the last step. To jointly learn the dense retriever, encoder, and decoder, two kinds of objective functions are used in this paper, i.e.~retrieval loss and bag-of-words loss. 

In summary, our contributions are as follows:

\begin{itemize}
\item Facing the inconsistency problem in unsupervised style transfer, we propose a novel transductive learning approach, which avoids learning a general rule but relies on specific samples to complete the style transfer process.

\item We design a \textbf{T}ran\textbf{S}ductive \textbf{S}tyle \textbf{T}ransfer (TSST) model, which employs a retriever to involve highly related samples to guide the learning of the target style. %Sparse and dense retrieval functions are adopted, and we build an end-to-end training of the dense retriever, encoder, and decode.

\item Experiments on two benchmark datasets show that TSST alleviates the inconsistency problem and achieves competitive results against traditional baselines. Our code is available at \url{https://github.com/xiaofei05/TSST}.
\end{itemize}

\section{Related Work}
~\label{sec:relatedwork}
% 是不是叫 embedding controled method, generator controled method, discriminator controled method 更合理啊？

Previous unsupervised text style transfer methods can be divided into three categories according to the way they control the text style, e.g. embedding based method, decoder based method, and classifier based method. 

The embedding based methods assign a separated embedding for each style to control the style of generated text. Early work tries to disentangle the content and style in the text. They first implicitly eliminate the original style information from the text representation using adversarial training ~\citep{shen2017style, fu2018style, john2019disentangled} or explicitly delete style-related words ~\citep{li2018delete, sudhakar2019transforming, wu2019mask, malmi2020unsupervised}. Then, decode or rewrite the style-independent content with the target style embedding. As a complete disentanglement is unreachable and damaged the fluency of the text, recent approaches ~\citep{lample2018multiple, dai2019style, yi2020text, zhou2020exploring, he2020probabilistic} directly feed original text representation and a separated learned style embedding to a stronger generator, e.g., the attention-based sequence-to-sequence model or Transformer to obtain the style transferred text. 
%recent approaches ~\citep{lample2018multiple, dai2019style, yi2020text, he2020probabilistic} abandoned the disentanglement of style and content. Although a style embedding was still fed to the decoder to control the style, they could adopt stronger generators, such as the attention-based sequence-to-sequence model and Transformer ~\citep{vaswani2017attention}, because there was no limits of having to get a representation of the input sentence. 

% 这里转折到disentangle的做法，但是区别讲的时候好像有点不清楚，我理解的是不去从原始句子中拆解出来Content和Style，而是直接输入一个Style来控制或者改写，而不管原始句子的Style？

The decoder based methods build a decoder for each style or transfer direction, where the style is implicitly represented as the parameters in the corresponding decoder. The former schema built an independent decoder for each style, which first disentangled the style-irrelevant content from the text and then applied the corresponding decoder to generate sentences with the target style \citep{fu2018style, xu2018unpaired, prabhumoye2018style, krishna2020reformulating}. The latter built for each transfer direction, which often regarded style transfer as a translation task ~\citep{zhang2018style, gong2019reinforcement, luo2019dual, jin2019imat, wu2019hierarchical, li2020dgst}. 
This paradigm reduced the complexity of the learning of style to a certain extent but consumed more resources. It is worth mentioning that the boundaries of embedding controlled and decoder controlled methods are sometimes not very clear, and many studies ~\citep{fu2018style, he2020probabilistic} consider that they are alternative.

The classifier based methods convert the style by manipulating the latent representation of the text according to a pre-trained classifier. \citet{wang2019controllable} and \citet{liu2020revision} mapped the input sentence into a latent representation, and trained classifiers on this latent space. The latent representation would be edited based on the gradients of the classifier until the predicted style changed; after that, the decoder took the modified representation to generate the desired style sentence. \citet{mai2020plug} further expanded this framework to a plug and play scene. Although it has remarkable style accuracy, this method can hardly guarantee content preservation due to the concrete output sharply changes with the latent representation.

We can see that all of these existing methods belong to the inductive learning approach, because they aim to learn a general style transfer rule from the training data, and then apply the rule to the test cases. Due to the lack of parallel corpus for supervision, this inductive learning approach fails to learn an accurate style representation application for various contexts, and may cause some severe inconsistency problems as illustrated before.

\begin{figure*}
\centering %使插入的图片居中显示
\includegraphics[scale=0.65]{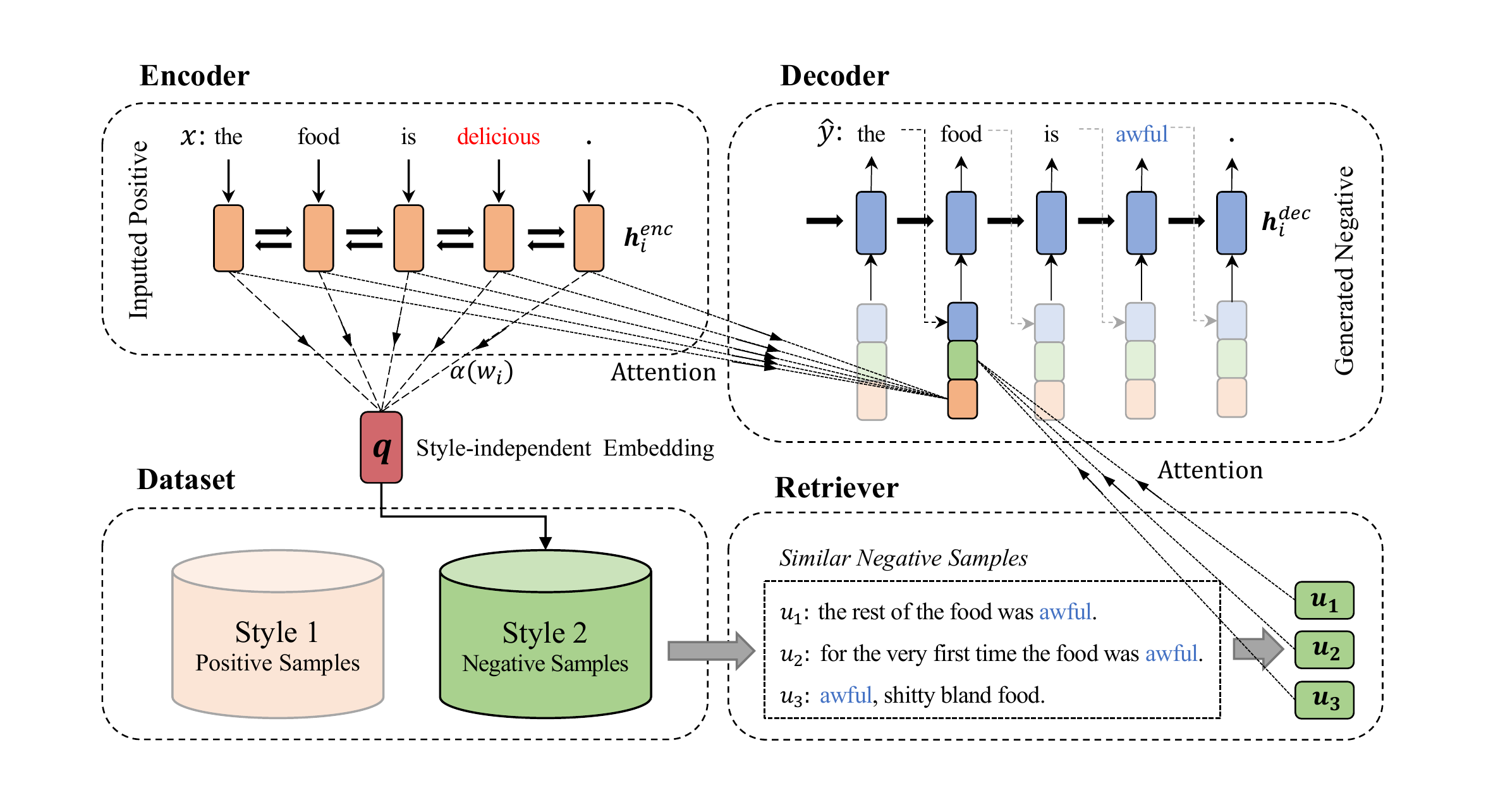}
\caption{\label{fig:model} Illustration of our TSST model for binary style transfer as an example.} %插入图片的标题，一般放在图片的下方，放在表格的上方
\end{figure*}

\section{Transductive Style Transfer}
Firstly, we introduce some notations. Consider the unsupervised text style transfer task with $M$ styles, its training set is composed of $M$ single-style subsets $\{D_i\}_{i=1}^M$. For an arbitrary input text $x$ in a subset and the target style $s_j$, the goal of text style transfer is to generate a new sentence $y$ which represents the style $s_j$ while keeping the style-independent content of $x$ as much as possible.

To tackle the aforementioned inconsistency problem, we propose to utilize transductive learning to obtain a context-aware style representation for the style transfer process. Specifically, our proposed transductive style transfer(TSST) model consists of three modules, encoder, retriever, and decoder, as described in Figure~\ref{fig:model}.

\subsection{Encoder} 
The goal of the encoder is to map the input sentence into hidden representations, to facilitate the following retrieval and generation process. Given the input sentence $x = (w_1, w_2, \dots, w_n)$, the output of the encoder is a sequence of hidden states,
\begin{equation} \label{Eq:encoder}
    \bm{H}^{enc} = \textrm{Encoder}(x),
\end{equation}
where $\bm{H}^{enc} = [\bm{h}^{enc}_1, \bm{h}^{enc}_2, \dots, \bm{h}^{enc}_n]^\mathrm{T} \in \mathbb{R}^{n\times d}$, and $d$ is the dimension of the hidden state. Please note that the encoder in our model is very general, and different encoding techniques can be used. Specially, we employ a bidirectional LSTM in our experiments.

\subsection{Retriever}
The retriever module is introduced to involve the top-$K$ relevant texts in the target style training subset, to facilitate the transductive learning process. In this paper, we adopt both sparse and dense retrieval functions in the retriever.

\paragraph{Sparse Retriever (BM25)} BM25~\citep{robertson2009probabilistic} is the most famous sparse retrieval function, which has been widely used in information retrieval.
\begin{equation}
	\label{BM25_F}
	\small
	\textrm{BM25}(\bm{q}, \bm{d}) = \sum_{w \in \bm{q}}  \frac{\textrm{IDF}(w) \cdot f(w,\bm{d}) \cdot (k_1+1)}{f(w,\bm{d}) + k_1 \cdot (1-b+\frac{b\cdot|\bm{d}|}{\text{avgdl}})}
\end{equation}
where $k_1$ and $b$ are the hyper parameters, $f(w,\mathbf{d})$ represents term frequency of $w$ in document $\mathbf{d}$, $\textrm{IDF}(w)$ represents inverse document frequency of $w$, $|\mathbf{d}|$ denotes the document length and $\text{avgdl}$ denotes the averaged document length.

After that, the retrieved $K$ texts $\{u_i\}_{i=1}^K$ are mapped to latent representations $\bm{U} = [\bm{u_1}, \bm{u_2}, \dots, \bm{u_K}]^\textrm{T} \in \mathbb{R}^{K\times d}$ by the above encoder, where $\bm{u_i}$ is the final hidden states for $u_i$.

%An alternative is the traditional sparse retriever BM25, which based on bag-of-words features. It also can get top-$K$ texts in the target style training subset. After that, these retrieved texts can be mapped in the same way to get latent representations $\bm{U}$.

\paragraph{Dense Retriever (MIPS)} A term-based sparse retrieval would have difficulty retrieving such a semantic context, which is essential for style transfer. Recently, dense retrieval methods and their efficient implementation of maximum inner product search (MIPS) \citep{shrivastava2014asymmetric, guo2016quantization, cai2021neural} have been proposed to capture the semantic. For a dense retriever, style-independent text embedding is crucial because we rely on this embedding to find similar samples in a different style subset. To this end, the style-independent embedding $\bm{q}(x)$ of the text $x$ can be represented as a linear combination of the hidden states $\bm{h}^{enc}_i$:
\begin{equation} \label{Eq:query}
    \small
    \bm{q}(x) = \textrm{Softmax}([\alpha(w_1), \dots, \alpha(w_n)]) \cdot \bm{H}^{enc},
\end{equation}
where the parameter $\alpha(w_i)$ is the weight for each word $w_i$. They are initialized as $\alpha(w_i) = 1-\sum_j |f_{s_j}(w_i)-1/M|$, where $f_{s_j}(w_i)$ is defined as the count of $w_i$ in the subset of style $s_j$ across its total count in the whole dataset. This initialization assigns a small weight to the discriminative words for each style, whose frequencies in different style subsets vary significantly. Consequently, the embeddings will focus on style-independent words, and help learn the style-independent embeddings.

%in order to conduct style-independent retrieval across different style subsets. 
%They are initialized according to the frequencies of the word $w_i$ across $M$ stylistic subsets. Specifically, $\alpha(w_i) = 2(1 - |f_{s_1}(w_i) - f_{s_2}(w_i)|)$ in the initialization phase, where $f_{s_1}(w_i)$ is the frequency of $w_i$ in the corpus of style $s_1$. The greater the frequencies difference, the smaller the weight of the word. In this way, the query embedding $\bm{q}$ is more inclined to words unrelated to the style.

Based on the text embeddings, the dense retrieval approach is used to retrieve the top-$K$ similar sentences in the target style training subset, where the cosine similarity function is used to measure the similarity. Note that computing text embeddings in the whole training set are time-consuming, so we pre-compute the text embeddings at the beginning and update them after certain training iterations, as inspired by ~\citet{guu2020realm}.

After that, the same encoder is employed to obtain the latent representations of the top-$K$ texts, i.e.~$\bm{U} = [\bm{u_1}, \bm{u_2}, \dots, \bm{u_K}]^\mathrm{T}$, as did in the sparse retriever.
%Next, we employ the same encoder to map the retrieved $K$ texts $\{u_i\}_{i=1}^K$ into latent representations $\bm{U} = [\bm{u_1}, \bm{u_2}, ..., \bm{u_K}] \in \mathbb{R}^{K\times d}$ , where $\bm{u_i}$ is the final hidden states for $u_i$.

% 公式里面的h和前面的对应关系是什么？前面用的是粗体的h，需要在图上面画出来。
\subsection{Decoder}
The decoder is used to generated the transferred text word by word. For each step, the inputs of the decoder are composed of three parts: 1) output of the previous step $\hat{y}_{t-1}$, 2) hidden states of $x$, i.e.~$\bm{c}_{t}^{h}$, and 3) latent representations of retrieved samples, i.e.~$\bm{c}_{t}^{u}$. The generated text $\hat{y} = \{\hat{y}_1, \hat{y}_2, \dots, \hat{y}_m\}$, is obtained by the following equations:
\begin{equation}
\begin{aligned}
    &\bm{h}^{dec}_{t} = \textrm{Decoder}(\hat{y}_{t-1}, \bm{c}_{t}^{h}, \bm{c}_{t}^{u}),\\
    &\textrm{P}(\hat{y}_{t}|\hat{y}_{<t}, \bm{H}^{enc}, \bm{U};\bm{\theta}) = \textrm{softmax}(\bm{o}_{\hat{y}_t} \bm{h}^{dec}_{t}), 
\end{aligned}
\end{equation}
where $\bm{o}_{\hat{y}_t}$ is the parameters to obtain the predicted probability on the word $\hat{y}_t$, and $\bm{c}_{t}^{h}, \bm{c}_{t}^{u}$ indicates the attended input sentence and retrieved samples, respectively. $\bm{c}_{t}^{h}$ and $\bm{c}_{t}^{u}$ are calculated by two attention modules with different parameters:
\begin{equation}
\begin{aligned}
&\bm{c}_{t}^{h} = \textrm{Attention}(\bm{h}^{dec}_{t-1}, \bm{H}^{enc}), \\
&\bm{c}_{t}^{u} = \textrm{Attention}(\bm{h}^{dec}_{t-1}\bm{W}_{h}, \bm{U}\bm{W}_{u}),
\end{aligned}
\end{equation}
where $\bm{W}_h \in \mathbb{R}^{d_{dec}\times d_{dec}}, \bm{W}_u \in \mathbb{R}^{d \times d}$ and $d_{dec}$ is the dimension of the decoder's hidden state. The attention module is standard~\citep{bahdanau2014neural}.\begin{equation}
\begin{aligned}
    &\textrm{Attention}(\bm{h}, \bm{H}) = \sum_{j=1}^n \frac{exp(e_{ij} )  \bm{H}_{j}}{\sum_{k=1}^n exp(e_{ik})}, \\
    & e_{ij} = \bm{v}^T \tanh(\bm{W}_d \bm{h} + \bm{W}_{e} \bm{H}_j),
\end{aligned}
\end{equation}
where $\bm{v}, \bm{W}_e, \bm{W}_d$ are parameters.

Similar to the encoder, various decoding techniques could be used in this step, and we adopt LSTM in our experiments.

\subsection{Learning Objectives}
Except for the three widely used losses in previous style transfer works, i.e.~the reconstruction loss, the cycle reconstruction loss, and the adversarial style loss, we also introduce two more losses related to the retriever, i.e.~the retrieval loss and the Bag-of-Word loss. Therefore, for a given input $x$, its style $s_i$ and the target style $s_j$, the learning objective function can be represented as: 
\begin{equation}
    \mathcal{L} = \mathcal{L}_{rec} + \mathcal{L}_{cyc} + \mathcal{L}_{adv} + \mathcal{L}_{ret} + \mathcal{L}_{bow},
\end{equation}
where $\mathcal{L}_{rec}$, $\mathcal{L}_{cyc}$, $\mathcal{L}_{adv}$, $\mathcal{L}_{ret}$ and $\mathcal{L}_{bow}$ denote the reconstruction loss, the cycle reconstruction loss, the adversarial style loss, the retrieval loss, and the bag-of-words loss, respectively. 

%The first three are often adopted in the embedding controlled methods, and $\mathcal{L}_{ret}$ and $\mathcal{L}_{bow}$ are newly proposed for the joint training of the dense retriever, encoder, and decoder.

\paragraph{Reconstruction Loss} According to previous works \citep{shen2017style, fu2018style, john2019disentangled}, this loss is used to capture the informative features for reconstructing itself:
\begin{equation}
    \mathcal{L}_{rec} = - \log P(x|x, s_i).
\end{equation}

\paragraph{Cycle Reconstruction Loss} Cycle consistency is usually included in the loss to improve the preservation of content \citep{lample2018multiple, dai2019style, yi2020text}. For the generated $\hat{y}$, the output of transferring back to the source style should be consistent with $x$ as much as possible:
\begin{gather}
    \mathcal{L}_{cyc} = - \log P(x|\hat{y}, s_i), \;\;
        \hat{y} = G(x, s_j),
\end{gather}
where $G$ is our TSST model. 

\paragraph{Adversarial Style Loss} 
%When the target style is not the same as the source style and there is no parallel corpus, 
If we only use the reconstruction and cycle construction losses, the model merely learns to copy the input to the output. So we employ adversarial training to build style supervision. Specifically, we utilize a classifier with $M+1$ classes as the discriminator $C$, similar to \citet{dai2019style} and \citet{yi2020text}. The first $M$ classes represent the real texts in the datasets, and the $(M+1)$-th class indicates generated fake texts. 
Since the generated text $\hat{y}$ is expected to be classified as the target style $s_j$, the adversarial style loss is defined as follows, and the negative gradient of the discriminator is employed to update the model. 
\begin{equation}
    \mathcal{L}_{adv} = - \log P_C(j|\hat{y}),
\end{equation}
As for the discriminator $C$, a loss function $\mathcal{L}_{C_1}$ is usually used in previous works \citet{dai2019style} and \citet{yi2020text}. In this paper, we also ask the discriminator to identify the style of the retrieved samples. Therefore, the loss function in our work can be written as:
%The discriminator is optimized by
\begin{equation}
\small
\begin{aligned}
    \mathcal{L}_{C} &= \mathcal{L}_{C1} + \mathcal{L}_{C2}, \\
    \mathcal{L}_{C1} &= - [\log P_C(i|x) + \log P_C(i|G(x, s_i)) \\ 
    &+ \log P_C(M+1|\hat{y})] , \\
    \mathcal{L}_{C2} &= - [\log P_C(j|Y^{x\rightarrow \hat{y}}) + \log P_C(i| Y^{\hat{y}\rightarrow x}) ] , \\
\end{aligned}
\end{equation}
where $Y^{x\rightarrow \hat{y}}$ and $Y^{\hat{y}\rightarrow x}$ denotes the retrieved samples in the transfer process from $x$ to $\hat{y}$ and from $\hat{y}$ to $x$, respectively.

To jointly learn the dense retriever with the other parameters and the target style representation from the retrieved samples, we introduce two additional losses to our objective.
%However, using the above three objective functions will cause two defects, 1) there is no direct supervision for our dense retriever, and 2) the target style expressions appeared in the retrieved samples is hard to be directly copied during the generation. Therefore, we propose two novel objective functions to tackle these defects.
%\paragraph{Cosine Similar Loss} 
\paragraph{Retrieval Loss} 
The retrieval loss is designed to capture the similarity between the style-independent embeddings of the input sentence and the corresponding transferred sentence:

%, which makes the embeddings of the input sentence and transferred sentence closer,

%From the definition of dense retrieval, the input sentence and transferred sentence are relevant, that is to say, their style-independent embeddings are similar. Therefore, we propose the retrieval loss, which makes the embeddings of the input sentence and transferred sentence closer,
\begin{equation}
    \mathcal{L}_{ret} = 1 - \cos(\bm{q}(x), \bm{q}(\hat{y})).
\end{equation}

%\paragraph{Content Loss} 
\paragraph{Bag-of-Words Loss}
This loss is proposed to encourage the generator to select some new words from the retrieved sentences, making our model pay more attention to the retrieved samples. In this way,  the style representation in the target sentence will be well adapted to the context. Let $\Omega$ denote a set of new words that appear in the retrieved samples other than the input sentence $x$, and the bag-of-words loss is defined as:
\begin{equation}
\begin{aligned}
    &\mathcal{L}_{bow} = \frac{1}{2}(\mathcal{L}_{bow}^{x\rightarrow\hat{y}} + \mathcal{L}_{bow}^{\hat{y}\rightarrow x}), \\
    &\mathcal{L}_{bow}^{x\rightarrow\hat{y}} = \frac{1}{\|\Omega\|}\sum\nolimits_{i=1}^{\|\hat{y}\|} \sum\nolimits_{w\in \Omega} \log  p_{\hat{y}_i}(w), \\
    &\mathcal{L}_{bow}^{\hat{y}\rightarrow x} = \frac{1}{\|\Omega'\|}\sum\nolimits_{i=1}^{\|x\|} \sum\nolimits_{w\in \Omega'} \log  p_{x_i}(w).
\end{aligned}
\end{equation}

Note that the Bag-of-Words loss is applied to both directions in the cycle reconstruction process.
% and it is different from BOW loss used in \citep{}...

\subsection{Discussion}
Please note that there are also some other works to involve a retrieval module to enhance the unsupervised style transfer, e.g.~\citet{li2018delete, sudhakar2019transforming, jin2019imat}. The differences between our TSSM and these works are listed as follows. 1) Our TSST model uses the retrieved samples to directly control the style, instead of using them as an external knowledge or pseudo-parallel corpus. 2) The retrieval module in our TSST can be trained in an end-to-end way (together with the encoder and decoder), to improve the style-independent text retrieval. 3) The retrieved samples influence the decoder word by word. In this way, the information of the retrieved samples will be fully exploited to learn a good context-aware style representation. It may bring no benefit if we just use the retrieved samples without these modifications, as shown in \citet{sudhakar2019transforming}. 
%Compared with previous work, our transductive style transfer(TSST) model is mainly innovated in two aspects. First, we utilize an additional retriever to obtain specific exaßmples with relevant content but different styles, which provide context-aware style representations to improve the consistency of transferred sentences. Second, two kinds of objective functions are newly designed to make the retrieved samples work and joint learning of the dense retriever, encoder, and decoder.

% It is not the first time that the retrieval has been used in the unsupervised style transfer. However, 

\section{Experiment}

In this section, we conduct experiments to study how well and why the proposed TSST model alleviates the inconsistency problem. Furthermore, a detailed ablation study is demonstrated to show each objective function's contribution to the overall performance.

\subsection{Datasets}
% 讲一下为什么是这两个数据集

The experiments are conducted on two well-known transfer tasks, sentiment transfer and formality transfer. The statistics of each datasets are shown in Table ~\ref{tab:datasets}.

\textbf{Yelp Dataset}\footnote{\url{http://bit.ly/2LHMUsl}.} is widely used as the benchmark for sentiment transfer. It is collected from restaurant and business reviews, with each text marked as positive or negative. The same pre-processing as \citep{li2018delete} is used in our experiment, and human references are also provided for the test set.

\textbf{GYAFC Dataset}\footnote{\url{https://github.com/raosudha89/GYAFC-corpus}.} denotes the Grammarly's Yahoo Answers Formality Corpus released by \citet{rao2018dear}, a typical benchmark for formality transfer. The GYAFC dataset contains formal and informal sentences in two different domains, \emph{Entertainment \& Music} and \emph{Family \& Relationships}. In this paper, we use the latter one because it is more popular in this area.

%\textbf{Sentiment Transfer (Yelp)} The Yelp dataset is widely evaluated for sentiment transfer. It is collected from restaurant and business reviews, each marked as positive or negative. We use the same precessing as ~\citep{li2018delete}, which also provides human references for the test set.

%\textbf{Formality Transfer (GYAFC)} We use Grammarly's Yahoo Answers Formality Corpus (GYAFC) released by ~\citet{rao2018dear} for formality transfer. The GYAFC dataset contains formal and informal sentences in two different domains. In this paper, we use the Family and Relationships domain. 

% The statistics of the two datasets are shown in Table ~\ref{tab:datasets}. Please note that we use training data as the retrieval corpus, whether in the training phase or in the test phase.

\begin{table}
\centering
\scalebox{0.9}{
    \begin{tabular}{ccccc}
    \hline
    Dataset            & Styles   & Train   & Dev   & Test  \\ \hline
    \multirow{2}{*}{Yelp}  & Positive & 266,041 & 2,000 & 500   \\
                           & Negative & 177,218 & 2,000 & 500   \\ \hline
    \multirow{2}{*}{GYAFC} & Formal   & 51,967  & 2,247 & 1,019 \\
                           & Informal & 51,967  & 2,788 & 1,332 \\ \hline
    \end{tabular}
}
\caption{\label{tab:datasets} Statistics of Yelp and GYAFC datasets.}
\vspace{-15 pt}
\end{table}

\begin{table*}
\centering
\resizebox{\textwidth}{!}{
\begin{tabular}{lccccclccccc}
\hline
\multirow{2}{*}{Model} & \multicolumn{5}{c}{Yelp}                                                &  & \multicolumn{5}{c}{GYAFC}                                               \\ \cline{2-6} \cline{8-12} 
                                & Acc$\uparrow$            & s-BLEU$\uparrow$      & r-BLEU$\uparrow$       & PPL$\downarrow$ & GM$\uparrow$             &  & Acc$\uparrow$            & s-BLEU$\uparrow$      & r-BLEU$\uparrow$       & PPL$\downarrow$ & GM$\uparrow$             \\ \hline
CrossAlign ~\citep{shen2017style}                      & 78.7          & 16.65          & 8.11           & 66  & 7.09           &  & 61.6          & 2.21           & 3.25           & \textbf{37}  & 3.32           \\
DRG ~\citep{li2018delete}   & 88.1 & 36.75 & 16.66 & 100 & 10.4 & & 58.2 & 31.57 & 21.88 & 103 & 9.65 \\
MultiDec ~\citep{fu2018style}                    & 45.4          & 40.07          & 15.07          & 188 & 8.51           &  & 24.5          & 16.08          & 11.95          & 151 & 5.54           \\
B-GST ~\citep{sudhakar2019transforming}   & 82.4 & 30.82 & 16.32 & 156 & 9.52 & & - & - & - & - & - \\
G-GST ~\citep{sudhakar2019transforming}   & 61.1 & 45.72 & 22.08 & 257 & 10.27 & & - & - & - & - & - \\

DualRL ~\citep{luo2019dual}               & 87.9          & 58.90 & 28.77          & 105 & 13.37          &  & 55.5         & 52.80          & 43.69          & 159 & 12.61         \\
StyTrans ~\citep{dai2019style}               & 86.0          & 59.46 & 27.32          & 154 & 12.90          &  & 60.3         & 61.15          & 43.95          & 168 & 13.33          \\
IMaT ~\citep{jin2019imat}   & \textbf{93.9} & 16.92 & 11.26 & \textbf{14} & 9.08 & &- & - & - & - & - \\
Revision ~\citep{liu2020revision}       &      90.6          &     13.23           &      7.93 &  21   &    7.45            &  &   39.6    &    27.64   &   20.83     &  66   &   8.59             \\
PFST ~\citep{he2020probabilistic}                           & 84.6          & 48.90          & 23.72          & 67  & 12.35          &  & 63.9          & 28.68          & 21.53          & 40  & 10.17          \\
StyIns ~\citep{yi2020text}                         & 90.9          & 53.10          & 26.09          & 110 & 12.80          &  & 69.9          & 61.87 & 47.80          & 140 & 14.32          \\ \hline
TSST-random                           & 88.8 & \textbf{59.64}         & 28.44 & 117 & 13.34 &  & \textbf{76.5}  & 62.84         &  50.13 &108   & 15.06  \\
TSST-sparse                           & 90.7 & 59.03          & 28.71 & 108 & 13.46 &  & 75.3 &  \textbf{64.03}         & 50.39 & 101  & \textbf{15.15}  \\
TSST-dense                           & 91.8 & 59.34          & \textbf{28.89} & 108 & \textbf{13.54} &  & 74.1  & 63.70           & \textbf{50.49}  & 103  & 15.06 \\
\hline
\end{tabular}
}
\caption{\label{automatic-evaluation}
Automatic evaluation results on Yelp and GYAFC dataset. Acc denotes the accuracy of generated samples judged by the pre-trained classifier. s-BLEU and r-BLEU stands for {\em self}-BLEU and {\em ref}-BLEU, respectively. \textbf{Bold} denotes the best value in terms of each metric.
\vspace{-15pt}
}
\end{table*}

\begin{table}
\centering
\scalebox{0.63}{
    \begin{tabular}{lccccccccc}
    \hline
    \multirow{2}{*}{Model} & \multicolumn{4}{c}{Yelp} &  & \multicolumn{4}{c}{GYAFC} \\ \cline{2-5} \cline{7-10} 
                           & Sty  & Cont & Flu & Cons &  & Sty  & Cont  & Flu & Cons \\ \hline
    DualRL                 &  3.70    &  4.34    &  \textbf{4.29}   &  1.38    &  &     1.84 &  2.92     & 3.11    &0.37      \\
    StyTrans               &  2.98    & 4.18     & 3.54    &    1.04  &  &  2.43    & 3.27      & 3.53    & 0.49     \\
    PFST                   & 3.21    & 3.69     & 4.18    & 1.26     &  &  1.97    &    2.28   &3.41     & 0.31     \\
    StyIns                 & 3.43  &     4.12 & 3.82     & 1.27     &  & 1.36     & 1.62      & 2.21    &  0.11    \\ \hline
    TSST-random          & 3.64  &  4.34    &  3.84   & 0.99     &  &  \textbf{2.84}    &  3.55     & 3.59    &  0.51    \\
    TSST-sparse          & 3.52    &  \textbf{4.42}    & 4.02    &    1.33  &  &  2.74    & 3.48      &   3.67  & \textbf{0.75}     \\
    TSST-dense           &  \textbf{3.85}    &    4.40  & 4.09    & \textbf{1.46}     &  &  2.70    & \textbf{3.59}      &  \textbf{3.69}   & 0.74     \\ \hline
    \end{tabular}
}
\caption{\label{human-evaluation}
Human evaluation results on Yelp and GYAFC dataset. The Krippendorff’s alpha of human rating on two datasets is 0.76 and 0.73, respectively, indicating acceptable inter-annotator agreement.
}
\vspace{-15 pt}
\end{table}

\subsection{Setups}
% 要按照三类来归纳Baselines
Our baselines cover three different kinds of inductive learning approaches, as described in Section ~\ref{sec:relatedwork}. For the embedding based method, we choose \textbf{CrossAlign} ~\citep{shen2017style}, ~\textbf{StyTrans} ~\citep{dai2019style}, ~\textbf{PFST} ~\citep{he2020probabilistic} and ~\textbf{StyIns} ~\citep{yi2020text}. For the generator based method, \textbf{MultiDecoder} ~\citep{fu2018style} and ~\textbf{DualRL} ~\citep{luo2019dual} are selected for comparisons. Then, ~\textbf{Revision} ~\citep{liu2020revision} is considered as the representative of the discriminator based methods. 
At last, we also compare our model with previous methods involved in retrieval, \textbf{DRG}~\citep{li2018delete}, ~\textbf{IMaT}~\citep{jin2019imat}, ~\textbf{B-GST} and ~\textbf{G-GST}~\citep{sudhakar2019transforming}.
All baselines (including generated results) are directly taken or implemented from their public source codes, so the detailed settings are omitted in our paper.

For our proposed TSST model, we employ the LSTM as our encoder and decoder to ensure the fairness of the experiment compared with previous methods. Following ~\citet{yi2020text}, we pre-train a forward LSTM language model in each dataset and use its parameters to initialize our encoder and decoder. Similar to ~\citet{yi2020text}, the discriminator is a CNN-based classifier with Spectral Normalization ~\citep{miyato2018spectral}, with the same word embeddings as the encoder. The word embedding size, hidden state size, and the number of retrieved samples $K$ are set to 256, 512, and 5, respectively. We exclude the trivial candidates the same as the input sentence in the retriever. The embeddings of all sentences for dense retrieval are updated every 200 steps.
To demonstrate the effectiveness of the sparse and dense retrievers, we compare them with a random sampling retriever, and the corresponding TSST model is denoted as TSST-random.

\subsection{Evaluation Metrics}

Previous works mainly focus on evaluating the style transfer methods from the following three aspects, i.e.~style transfer accuracy, content preservation, and sentence fluency. Consequently, different automatic evaluation measures such as accuracy, \emph{self}-BLEU, \emph{ref}-BLEU, and perplexity(PPL) are also used in the evaluation. However, all of these metrics cannot well evaluate how well a model alleviates the consistency problem, as we introduced before. So we introduce an additional human evaluation in our experiment.

%We evaluate the style transfer task from four aspects: , and collocation consistency. 

%which can not fully evaluate the consistent in generated sentence, as has shown in Figure~\ref{fig:introduction}. Therefore, we propose an additional collocation consistency metric in human evaluation.

%Following previous work ~\citep{fu2018style, luo2019dual, john2019disentangled, dai2019style, he2020probabilistic, yi2020text}, we evaluate the task from three aspects: style transfer accuracy, content preservation, and fluency. However, these metrics can not fully evaluate the consistent in generated sentence, for the inconsistent sample in Figure~\ref{fig:introduction} may have high scores in above three aspects. Therefore, we need to add a consistency metric in human evaluation to evaluate the effectiveness of the transductive learning.
%For a formality transfer case, the informal sentence is \emph{i hope u will answer me} and the transferred formal sentence is \emph{i hope it will answer me}. It will earn higher scores in these three aspects, but it fails to the consistency. Therefore, we add a consistency metric in human evaluation to evaluate the consistency of modified parts.

\textbf{Automatic Evaluation} To evaluate the style transfer accuracy, we first finetune a pre-trained BERT-based \citep{devlin2019bert} classifier on each dataset. The two classifiers achieve 98.6\% and 89.9\% accuracy on the test set of Yelp and GYAFC, respectively. Then these classfiers are used to predict the style label of the generated transferred sentences, and the classification accuracy acts as the style transfer accuracy. Both \emph{self}-BLEU and \emph{ref}-BLEU are used for content preservation evaluations. The former is the BLEU score between transferred sentences and source sentences, while the latter is between transferred sentences and human references. Following \citet{dai2019style} and \citet{yi2020text}, we train a 5-gram language model KenLM \citep{heafield2011kenlm} for each style to measure the lanuguage fluency by the perplexity (PPL) of transferred sentence. In addition, we report the geometric mean (GM) of Acc, \emph{self}-BLEU, \emph{ref}-BLEU and $\frac{1}{\log \rm{PPL}}$ as the overall performance.

\textbf{Human Evaluation} 
%Since automatic evaluation may not sufficiently reflect the performance of the style transfer, especially the collocation consistency. 
We recruit three annotators who have high-level language skills for human evaluation. We choose four models with the highest GM scores and three types of TSST models in this experiment. Following previous works \citep{dai2019style, yi2020text, liu2020revision}, we randomly selected 100 generated sentences (50 for each style) in the test set and the annotators are required to score sentences from 1 to 5, in terms of each aspect, i.e.~style transfer accuracy (Sty), content preservation (Cont), and sentence fluency (Flu), where 1 is the lowest and 5 is the highest. In addition, they need to evaluate the consistency of the style words and other contexts in each generated sentence. To make it more clear, consistency (Cons) focuses on judging whether the modified parts are consistent with the retained content, while the fluency only focuses on the grammatical errors. The consistency is rated from 0 to 2, where 0, 1, and 2 stands for inconsistent, unsure, and consistent, respectively.

\subsection{Experimental Results}

\begin{table*}[t]
\centering
\scalebox{0.75}{
    \begin{tabular}{c|l|l}
    \hline
    \textbf{Model}             & \multicolumn{1}{c}{\textbf{Negative to Positive}}            & \multicolumn{1}{|c}{\textbf{Positive to Negative}} \\ \hline
    input                      & it 's just \textcolor{blue}{too expensive} for what you get.                  & mustard beef ribs are \textcolor{red}{a must}.                    \\ \hline
    DualRL                     & it 's just \textcolor{red}{fun} for what you get.                            & mustard beef ribs are \textcolor{blue}{a mess}.                    \\
    StyTrans                   & it 's just \textcolor{blue}{too expensive} for what you get.                  & mustard beef ribs are \textcolor{blue}{a negative}.                \\
    StyIns                     & it 's just \textcolor{red}{very fun} for what you get.                       & mustard beef ribs were a total.                  \\
    PFST                       & it 's just \textcolor{red}{right} for what you get.                          & gross.                                           \\
    Revision                   & it 's \textcolor{blue}{too expensive} for what you get and it 's always \textcolor{red}{good}. & the beef ribs are a must do not have a bbq beef. \\ \hline
    TSST-dense               & it 's just \textcolor{red}{really \textbf{reasonable}} for what you get.              & mustard beef ribs are \textcolor{blue}{\textbf{a joke}}.                    \\
    \multirow{5}{*}{Retrieved Samples} & \; $u_1$ - the price is \textcolor{red}{\textbf{reasonable}} for what you get.            & \; $u_1$ - garlic mashed potatoes were \textcolor{blue}{\textbf{a joke}}.              \\
                               & \; $u_2$ - prices are \textcolor{red}{very \textbf{reasonable}} for what you get.                   & \; $u_2$ - all olive gardens are \textcolor{blue}{\textbf{a joke}}.                    \\
                               & \; $u_3$ - prices are \textcolor{red}{\textbf{reasonable}} for the qualiy of what you get.                   & \; $u_3$ - the ribs are \textcolor{blue}{over cooked}.                        \\
                               & \; $u_4$ - \textcolor{red}{good} food , a little expensive for what you get.                & \; $u_4$ - our food was \textcolor{blue}{barely edible}.               \\
                               & \; $u_5$ - definitely a \textcolor{red}{good} price for what you get.      & \; $u_5$ - cold grits are \textcolor{blue}{n't a treat}.                      \\ \hline
    \end{tabular}
}
\caption{\label{case-study}Case Study. \textcolor{red}{Red} denotes \emph{positive} expressions, \textcolor{blue}{blue} denotes \emph{negative} expressions, and \textbf{bold} denotes the expression taken from retrieved samples.}
\vspace{-15pt}
\end{table*}

\textbf{Automatic Evaluation} As listed in Table~\ref{automatic-evaluation}, we can see that our transductive learning models significantly improve the overall performance on both datasets, as compared with the inductive learning baselines. 
As for the four specific evaluation measures, our model achieves better results in terms of three of them, i.e.~Acc for style transfer accuracy, s-BLEU, and r-BLEU for content preservation, and all our models are able to generate sentences with relatively low perplexity. Though previous models achieve the best on a single metric, a significant drawback can be found on another metric. 
For the TSST-random, although target-style relevant words in the random samples are not consistent with the input content, the few modifications and the use of target-style strong relevant words will still make the random baseline achieve the high automatic metrics. Thus we need human evaluation.

% It is worth mentioning that TSST-random also achieves good performance. In fact, the improvement of performance not only comes from the retrieval component but also the well-designed loss function, i.e., Bag-of-Words loss. It will impel the model to “pick” the target-style relevant words and use them to transfer the input sentence to the target style at the least modification cost. Although target-style relevant words in the random samples are not consistent with the input content, the few modifications and the use of target-style strong relevant words will still make the random baseline achieve the high automatic metrics.

%However, automatic evaluation is hard to reflect the quality of transferred text, especially the inconsistency problem focused in this paper, thus we should draw more attention to human evaluation metrics.

\textbf{Human Evaluation} Results are shown in Table ~\ref{human-evaluation}. Firstly, the comparison results are consistent with the automatic evaluation results in terms of both style accuracy, content preservation, and fluency, indicating the reliability of our human evaluation. More importantly, our TSST-sparse and TSST-dense models achieve the highest consistency score, as compared to other baselines, showing the superiority of our transductive learning approach in tackling the inconsistency problem. In contrast, the consistency will become worse if the retrieved samples are not related to the input, as shown in TSST-random, which further demonstrate the soundness of our approach. 
Comparing TSST-sparse and TSST-dense, we can see that joint learning retriever yields better consistency results, which is accordant with previous studies. 

%other three traditional metrics, which can draw a same conclusion as in automatic evaluation.

\textbf{Case Study} To better understand what transductive learning bring to text style transfer task, Table ~\ref{case-study} shows some transferred examples from the Yelp test set. For the first example to transfer from negative to positive, DualRL and StyIns are able to capture the style transfer but with inappropriate expressions, e.g.~`\emph{fun}' or `\emph{very fun}'. StyTrans and Revision fail to do the style transfer, either just copy or use negative expressions. Our TSST-dense model produces perfect results by using `reasonable' as the transferred style expression, learned from the retrieved examples, as shown in the table. Similar results are observed for the second example to transfer positive to negative. More importantly, although there is no retrieved examples containing exactly the same phrase `\emph{mustard beef ribs}', our model still capture the negative pattern like `\emph{[food] is a joke}' to complete the style transfer process.

%to generalizes `\emph{mustard beef ribs}' to the semantic food category to guide the generation. 

%\subsection{Ablation Studies}
\begin{table}
\centering
\scalebox{0.75}{
    \begin{tabular}{cccccc}
    \hline
    \multicolumn{1}{c}{Model}       & Acc$\uparrow$           & s-BLEU$\uparrow$         & r-BLEU$\uparrow$         & PPL$\downarrow$          & GM$\uparrow$             \\ \hline
    \multicolumn{1}{l}{TSST-dense} & 91.8          & 59.34          & \textbf{28.89} & \textbf{108} & \textbf{13.54} \\ \hline
    $-$ $\mathcal{L}_{rec}$                    & 52.7 & 0.40          & 0.37          & N/A          & 1.01          \\
    $-$ $\mathcal{L}_{adv}$                    & 81.3 & 18.96          & 8.48          & N/A          & 6.67          \\
    $-$ $\mathcal{L}_{cyc}$                    & 89.0 & 52.07          & 24.71          & 134          & 12.37          \\
    $-$ $\mathcal{L}_{ret}$                    & 89.6          & 59.36         & 28.65          & 110          & 13.42          \\
    $-$ $\mathcal{L}_{bow}$                    & 89.3          & 55.26 & 26.43          & 121          & 12.84          \\  \hline
    $-$ $\mathcal{L}_{C1}$            & \textbf{95.6}          & 7.6          & 3.9           & N/A          & 4.47          \\ 
    $-$ $\mathcal{L}_{C2}$            & 87.4          & \textbf{61.41}          & 28.58         & 112          & 13.43          \\ \hline
    \end{tabular}
}
\caption{\label{ablation-study}Ablation study on the Yelp dataset, where `$-$' means removing one of the loss terms in the objective functions, and N/A is a very large value.}
\vspace{-15pt}
\end{table}

\textbf{Ablation Study} To study the role of each loss function in the objective function, we remove them one at a time and train the model from scratch. Due to the high cost of the human evaluation, we only report the automatic results on the Yelp dataset, as shown in Table~\ref{ablation-study}. 
We can see that each loss contributes to the performance of the model, and their combination performs the best. 
%Without one of the vanilla losses $\mathcal{L}_{rec}, \mathcal{L}_{adv}$ or $\mathcal{L}_{C_1}$, the model can not generate meaningful sentences. After the $\mathcal{L}_{ret}$ removed, the dense retriever is reduced to a heuristic retriever, resulting in the decrease of the overall performance. If the $\mathcal{L}_{bow}$ is disabled, the model does not have to use the expressions in retrieved samples, which causes more modifications without the guidance.

\begin{figure}
\centering %使插入的图片居中显示
\includegraphics[scale=0.55]{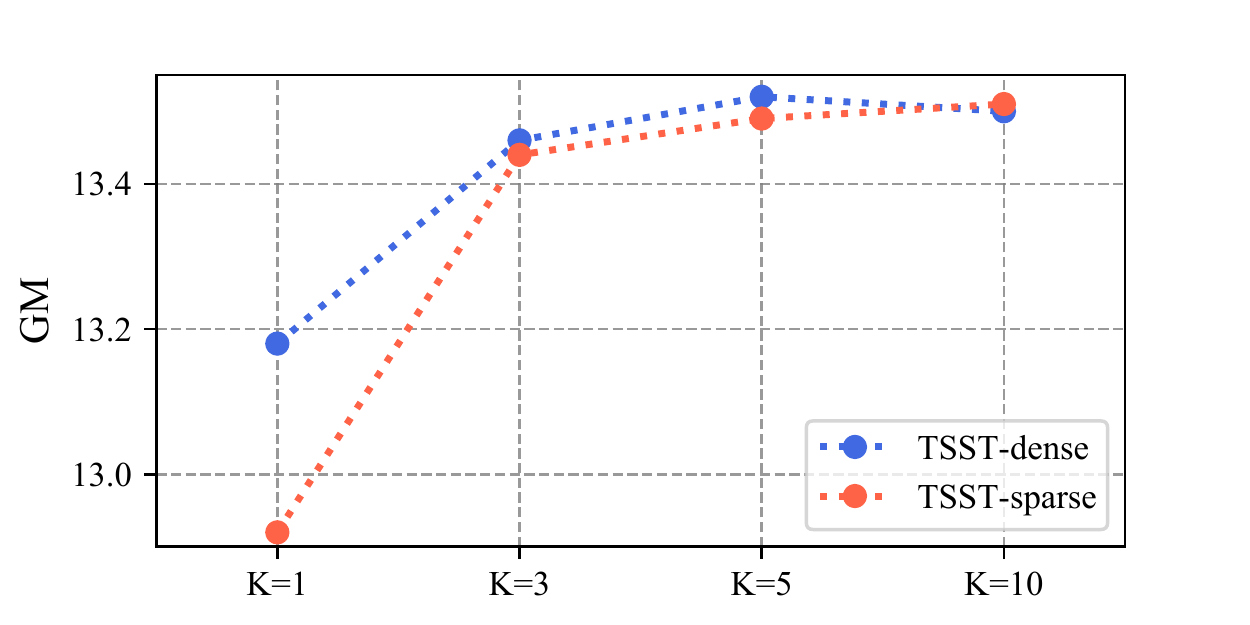}
\caption{\label{param-study} The influence of the number of retrieved samples $K$ on the Yelp dataset.  } %插入图片的标题，一般放在图片的下方，放在表格的上方
\vspace{-15pt}
\end{figure}

We also conduct experiments to explore the influence of different numbers of retrieved samples, as shown in Figure ~\ref{param-study}. Specifically, we test four different values, i.e.~$K=1,3,5,10$. We can see that the overall performance of the model gradually increases with the increase of $K$, and become stable for 3, 5, and 10. For the sake of both effectiveness and efficiency, we set $K=5$ in our experiments.

\section{Conclusions and Future Work}
Previous style transfer models are mainly based on inductive learning approach, thus suffer from the inconsistent style expression problem with lack of the parallel corpus as supervisions. To tackle this problem, we propose a novel transductive learning approach for unsupervised text style transfer. The key idea of our TSST model is to learn context-aware style expressions via retrieved samples from the target style datasets. Experimental results on two typical style transfer tasks show that TSST significantly improves the performances in terms of both automatic and human evaluation.
%Retrieval module can be jointly trained with encoder and decoder, by applying two new designed objective functions.
%In this way, we avoid the inductive learning of the style and alleviate the inconsistency problem. 
%Based on the embedding-based inductive framework which often involves a style embedding to transfer the style of the input sentence during the reconstruction, we abandon the controlling way of style embeddings and employ a retriever to provide style representations relevant to the input sentence.

%We use sparse (BM25) and dense retrieval functions (DPR) respectively, and design two objective functions to combine the dense retriever, encoder, and decoder for joint learning. Both automatic and human evaluation results on two benchmark datasets show the effectiveness of our method.

Our proposed transductive learning approach is very general, and this work mainly focus on the embedding-based methods. In future, we plan to extend our approach to other methods, such as decoder-based and discriminator-based methods. Moreover, we will try more powerful retrieval methods, such as DPR ~\citep{karpukhin2020dense}.

%Thus, it is worth to explore how the decoder and discriminator can be customized through specific samples.

\section*{Ethical Considerations}
We honor and support the ACL code of Ethics. The paper focuses on style transfer, which aims to change the style of the text while preserving the semantic content. We recognize the style transfer methods may be misused to generate misinformation, e.g. fake customer reviews. However, the style transfer methods can also provide strong support for mitigating harmful biases in online information, e.g. the transfer from offensive to non-offensive ~\citep{dos2018fighting, tran2020towards} and from biased to neutral ~\citep{pryzant2020automatically}. Overall, it is still meaningful to continue research into this work on the basis of predecessors. Simultaneously, the datasets we used in this paper are all from previously published works and do not involve privacy or ethical issues. 

\section*{Acknowledgements}

This work was supported by National Natural Science Foundation of China (NSFC) under Grants No. 61906180, No. 61773362 and No. 91746301, National Key R\&D Program of China under Grants 2020AAA0105200 and the Tencent AI Lab Rhino-Bird Focused Research Program (No. JR202033).  

\bibliography{anthology,custom}
\bibliographystyle{acl_natbib}

\end{document}